\documentclass[conference]{IEEEtran}
\usepackage{array}
\usepackage{verbatim}
\usepackage{amsthm}
\usepackage{amsmath,amssymb}
\usepackage{graphicx}
\usepackage{url}
\usepackage{algorithm2e}

\ifCLASSOPTIONcompsoc
  \usepackage[caption=false,font=normalsize,labelfont=sf,textfont=sf]{subfig}
\else
  \usepackage[caption=false,font=footnotesize]{subfig}
\fi

\newtheorem{example}{Example}

\begin{document}
\title{Enacting textual entailment and ontologies for automated essay grading in chemical domain}

\author{\IEEEauthorblockN{Adrian Groza and Roxana Szabo\\
\IEEEauthorblockA{Intelligent System Group,\\Department of Computer Science,\\ Technical University of Cluj-Napoca, Romania\\
Adrian.Groza@cs.utcluj.ro, Roxana.Szabo@cs-gw.utcluj.ro \\ 
}
}
}

\maketitle

\begin{abstract}
We propose a system for automated essay grading using ontologies and textual entailment.
The process of textual entailment is guided by hypotheses, which are extracted from
a domain ontology. Textual entailment checks if the truth of the hypothesis follows from a given text.
We enact textual entailment to compare students
answer to a model answer obtained from ontology. 
We validated the solution against various essays written by students in the chemistry domain.
\end{abstract}


\section{Introduction}

Assessment is an essential part of the learning process, especially in formative learning settings. 
In the current context of massive open online courses (MOOC), assessment is challenging as it aims to ensure consistency, reliability and do not favor one person against another. 
In formative assessment the problem of workload and timely results  is even greater, as the task is carried out more frequently while the interpretation of one human marker differs from another.

While essay questions are advantageous to student learning and assessment there are obvious disadvantages for the instructor. 
Grading of essay and discussion questions is time consuming even with the help of teaching assistants. 
Automated essay grading~\cite{shermis2013handbook} aims at automatically assigning a grade to a student’s essay by means of various features.
Since the argument structure is crucial for evaluating essay quality, persuasive essays are extensively studied~\cite{stab2014identifying}.
By automatically identifying arguments, the evaluator is be able to inspect the essay's plausibility.
We argue that information technology is able to assist and support teachers in these challenges.

Our research hyphotesis relies on the correlation between textual entailment~\cite{androutsopoulos2010survey} and answer correctness. 
In a typical answer assessment scenario, we expect a correct answer to entail the reference answer. 
However a student may wish to skip the details already mentioned in the question. 
Hence, the problem is whether the answer, along with the question, entail the reference answer. 

Let the question be $a$, student answer  be $s$ and the reference answer be $r$. 
Correctness means $a \wedge s  \Rightarrow r$ and contradiction means $s \wedge a \Rightarrow \neg r$. 
We propose the usage of recognizing textual entailment (RTE) along with shallow text features to train on the system dataset and testing it on the test dataset provided. 
The evaluation metrics used will be according to the Coh-Metrix system~\cite{graesser2011coh}. 
The final grade will be obtained from the first  grade computed from the comparison between the student’s answer and  the hypotheses generated from the model ontology, and the second grade 
that will be obtained from evaluation of the metrics. 

The rest of the paper is organized as follows: 
Section~\ref{sec:architecture} presents the system architecture and the NLP tools enacted. 
Section~\ref{sec:scenario} details a running scenario on students essays in the chemical domain.
Section~\ref{sec:related} discusses related work, while section~\ref{sec:con} concludes the paper.

\section{System architecture}
\label{sec:architecture}

\begin{figure}
\begin{center}
\includegraphics[scale=0.45]{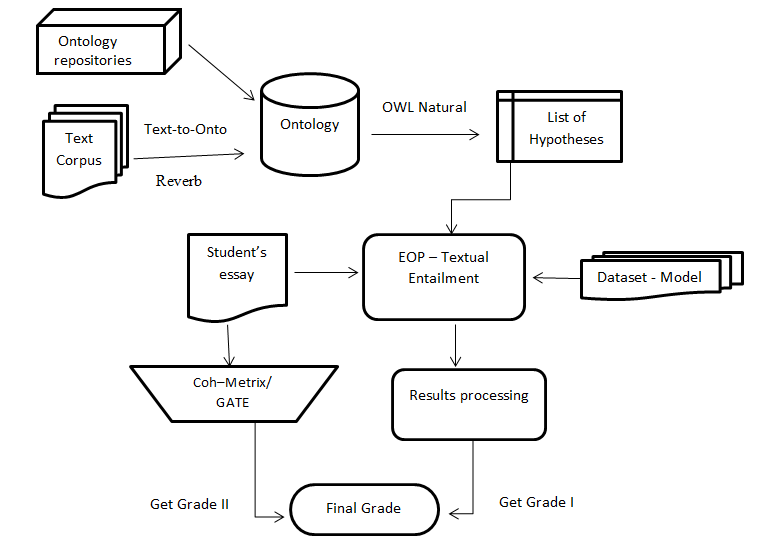}
\end{center}
\caption{System architecture.}
\label{fig:architecture}
\end{figure}

The proposed ontology-based essay grading system 
(OntoEG\footnote{The tool is available at http://cs-gw.utcluj.ro/$\sim$adrian/tools/ontoeg}) consists of a set of integrated natural language tools (see Fig.~\ref{fig:architecture}). 
The system is structured on layers.
The first layer contains the Text2Onto tool~\cite{cimiano2005text2onto}, used to obtain a consistent ontology from a corpus of high ranked or relevant essays in a given domain.
The second layer exploits the OWLNatural service~\cite{galanis2007generating}, to generate natural text from a selected ontology. 
We organise the text generated by OWLNatural as a set of hypotheses. 

In the second layer, textual entailment is used to analyse the domain hypotheses on the available essays. 
The EOP system~\cite{magnini2014excitement} is trained using a set of pairs $\langle Text, Hyphotesis \rangle$.  
For experiments, we created a data set in the chemical domain containing 100 pairs of text/hypothesis divided into 50"\%" of entailment pairs and 
50"\%" of non-entailment pairs.
The text is represented by the student's essay to be reviewed. 
The hypotheses are generated from a domain ontology and filtered by the teacher. 
Based on the model generated after training, the EOP system computes the confidence of hypotheses entailment within the text. 
This confidence constitutes the basis for grading the essay.

Automatic grading includes also various readability metrics.
For this step, we use GATE tool for natural language processing~\cite{cunningham2002gate}, 
to get the number of tokens or number of sentences from text. 
We integrate Coh-Metrix service~\cite{graesser2004coh} to compute 
various cohesion and coherence metrics for written texts. 

The system components and the main workflow appear in Fig.~\ref{fig:architecture}.
The following four components are detailed: 
(1) developing the domain ontology, 
(2) generating hypothesis from ontology,
(3) textual entailment methods, and  
(3) natural language processing of essays.

\subsection{Developing the domain ontology}

We assume that the professor provides a corpus of relevant documents in the domain of interest. 
Our approach is to automatically generate a domain ontology from this corpus. 
For this task, we rely on Text2Onto framework for ontology learning from textual resources. 

Three main features distinguish Text2Onto: 
Firstly, learned knowledge is represented at a meta-level Probabilistic Ontology Model (POM).
Secondly, user interaction is a core aspect of Text2Onto and the fact that the system calculates a confidence for each learned object
allows to design visualizations of the POM. 
Thirdly, by incorporating strategies for data-driven change discovery, we avoid processing the whole corpus from 
scratch each time it changes. Instead, POM selectively updates itself, according to the corpus changes only. 
Besides increasing efficiency, this solution allows to trace the evolution of the ontology with respect to the changes in the underlying corpus.

Text2Onto combines machine learning approaches with basic linguistic processing such as tokenization or lemmatizing and shallow parsing~\cite{cimiano2005text2onto}.
Since it is based on the GATE framework it is very flexible with respect to the set of linguistic algorithms used. 
Another benefit of using GATE is the seamless integration of JAPE rules which provides finite state transduction over annotations based on regular expressions.

The main workflow of ontology generation consists of: preprocessing, execution of algorithms, combining results. 
During preprocessing, Text2Onto calls GATE applications to tokenize the document, split sentences, tag Part of Speech and match JAPE rules.
GATE creates indexes for the document and the result of this is obtained as an AnnotationSet. 
In the next step, Text2Onto executes the applied algorithms in a pre-specified order: i) concept, ii) instance, iii) similarity, iv) subclass-of, v) instance-of, vi) relation and vii) subtopic-of. 
The basic heuristic employed in Text2Onto to extract concepts and instances is that nouns represent concepts and proper nouns are instances. 
If more than one algorithm is applied for each category, then the final relevance value is computed based on the
selected combiner strategies~\cite{Cunningham2002}.

Given a student essay, we need to analyse its content against the domain ontology available from the previous step.
We rely on ReVerb~\cite{fader2011identifying} to extract triplets from the student essay. 

ReVerb is designed for Web-scale information extraction, where the target relations cannot be specified in advance and speed is important. 
ReVerb first identifies relation phrases that satisfy the syntactic and lexical constraints, and then finds a pair of NP arguments for 
each identified relation phrase. 
A confidence score is assigned to the resulting extractions using a logistic regression classifier. 

This algorithm differs in three important ways from previous Open IE systems like TextRunner~\cite{yates2007textrunner}.
Firstly, the relation phrase is identified “holistically” rather than word-by-word.
Secondly, potential phrases are filtered based on statistics over a large corpus (the implementation of our lexical constraint). 
Finally, ReVerb is ``relation first'' rather than ``arguments first'', which avoids a common error like confusing a noun in the relation phrase for an argument, (e.g. the noun ``deal'' in ``made a deal with'')~\cite{fader2011identifying}.

Given an input sentence $s$, ReVerb uses extraction algorithm~\ref{alg:extraction}.
In the second part of the algorithm, for each relation phrase $r$ identified in Step 1, find the nearest noun phrase $x$ to the left of $r$ in sentence $s$ such that $x$ is not a relative pronoun or the existential “there”. 
Then, the algorithm finds the nearest noun phrase $y$ to the right of $r$ in sentence $s$. 
If such an $(x, y)$ pair could be found, the tuple $\langle x, r, y\rangle$ is returned.

\begin{algorithm}
\caption{Relation extraction algorithm.}
\SetKwInOut{Input}{Input}
\Input{
$s$ - sentence\;} 
\ForEach{verb $v\in s$}{
 find the longest sequence of words $rv$ such that:\\
(1) $rv$  starts at $v$, \\
(2) $rv$ satisfies the syntactic constraint, and\\
(3) $rv$ satisfies the lexical constraint}
\If{$\exists$ pair of matches adjacent or overlap in $s$}{merge them into a single match\;}  
 \ForEach{relation phrase $r$}{find the nearest noun phrase $x$ to the left of $r \in s$ such that $x \notin RelativePronoun \cup \{There\}$,\\ 
Find the nearest noun phrase $y$ to the right of $r \in s$.} 
 return $\langle x, r, y \rangle$.
\label{alg:extraction}
\end{algorithm}

Open Information Extraction (IE) is the task of extracting assertions from massive corpora without requiring a pre-specified vocabulary. 
ReVerb takes raw text as input, and outputs triplets $\langle argument1,\ relationPhrase, argument2 \rangle$, as illustrated in example~\ref{ex:triplet}.

\begin{example}[Triplets extraction with ReVerb]
Given the sentence: ``Vitamin D	is toxic in large amounts.'', 
the extracted triple is:
$\langle vitamin\ d,\ be\ toxic\ in,\ large\ amount\rangle$. 
For ``Bananas are an excellent source of potassium", ReVerb extracts the triple
$\langle bananas,\ be\ source\ of,\ potassium\rangle$.
\label{ex:triplet}
\end{example}  

\subsection{Generating hyphothesis}

We use OWLNatural to generate natural language hypotheses from a domain ontology. 
OWLNatural is natural language generation engine that produces descriptions of individuals and classes in 
English and Greek from ontologies that have been annotated with linguistic and user modeling information expressed in RDF.
The OWL verbalizer takes its input in OWL syntax and produces an output in a fragment of Attempto Controlled English (ACE)~\cite{fuchs2008attempto}.

\begin{figure}

Every OrganicSulfurCompound $\equiv$ OrganicCompound $\sqcap$ $\exists$ hasPart . OrganicSulfurGroup.\\

\begin{tabular}{lp{7cm}}\hline
$H_1$ & Every OrganicSulfurCompound is an OrganicCompound that hasPart an OrganicSulfurGroup.\\
$H_2$ & Every OrganicCompound that hasPart an OrganicSulfurGroup is an OrganicSulfurCompound.\\
\hline
\end{tabular}

\caption{Generating natural language hypothesis from chemical ontology.}
\label{fig:verbalizer}
\end{figure}

\subsection{Enacting textual entailment} 

Textual entailment (TE) is a directional relation between text fragments. The relation holds whenever the truth of one text fragment follows 
from another text. Given two text fragments, one named Text ($T$) - the entailing and the other named Hypothesis ($H$) - the entailed. 

Recognizing Textual Entailment (RTE) has been proposed~\cite{dagan2006pascal} as a generic task that captures major semantic inference needs across many natural language processing applications. 
The Recognizing Textual Entailment task consists in recognizing whether the Hypothesis 
can be inferred from the Text. 
We use a graduated definition of entailment: $T$ entails $H$ ($T \Rightarrow H$) if, typically, a human reading $T$ would 
infer that $H$ is most likely true. 
Positive entailment is illustrated in example~\ref{ex:positive-entailment}.

\begin{example}[Positive entailment]

$ $

\begin{tabular}{lp{7cm}}
$T$: &  {\it In chemical reactions with metals, nonmetals gain electrons to form negative ions. }\\
$H$: & {\it The nonmetals become negative ions.}\\
\end{tabular}
\label{ex:positive-entailment}
\end{example}

The correctness means that the text entails the hypothesis, so we obtain the positive entailment. 
The contradiction means that the text does not entail the hypothesis, and there are an negative entailment.
An example of a negative TE (text contradicts hypothesis) is illustrated by example~\ref{ex:negative-entailment}.

\begin{example}[Negative entailment]$ $\\
\begin{tabular}{lp{7cm}}
$T$: & {\it Nonmetallic elements also react with other nonmetals, in this case forming molecular compounds. }\\
$H$: &  {\it Metals react with nonmetals in order to form ions.}\\
\end{tabular}
\label{ex:negative-entailment}
\end{example}

An example of a non-TE (text does not entail nor contradict) is illustrated by example~\ref{ex:non-entailment}.

\begin{example}[Non entailment]

$ $

\begin{tabular}{lp{7cm}} 
$T$: & {\it A chemical reaction is one in which the organization of the atoms is altered. }\\
$H$: &   {\it The burning of methane is a chemical reaction is in the presence of oxygen.  }\\
\end{tabular}
\label{ex:non-entailment}
\end{example}

The Excitement Open Platform (EOP) is a generic architecture for textual inference in multiple languages. 
The platform includes state-of-art algorithms, a large number of knowledge resources, and facilities for experimenting. The input consists of the text $T$ and hypothesis $H$. 
The output is an entailment judgment, either ”Entailment” if $T$ entails $H$, or ”NonEntailment” if the relation does not hold. 
A confidence score for the decision is also returned in both cases. 

The overall structure consists of two main parts: Linguistic Analysis Pipeline and Entailment Core.
The Linguistic Analysis Pipeline (LAP) is a series of linguistic annotation components range from tokenization to part of speech tagging, chinking, Named Entity Recognition and parsing.
Entailment Core consists of Entailment Decision Algorithms (EDAs) and more subordinate components. An EDA takes an entailment decision while
components provide static and dynamic information for the EDA.
The Entailment Decision Algorithm (EDA) computes an entailment decision for a given Text/Hypothesis pair, and can use components that 
provide standardized algorithms or knowledge resources. Currently, the EOP ships with three EDAs each following a different approach: 
transformation-based, edit-distance based, and classification based. 
Scoring Components accept a Text/Hypothesis pair as an input, and return a vector of scores.
Distance Components that can produce normalized and unnormalized distance/similarity values in addition to the score vector.
Annotation Components can be used to add different annotations to the Text/Hypothesis pairs. 
Syntactic Knowledge Components capture entailment relationships between syntactic and lexical-syntactic expressions. 

Knowledge is needed to recognize cases where $T$ and $H$ use different textual expressions (words, phrases) while preserving entailment (e.g., home $\rightarrow$ house, Hawaii $\rightarrow$ America, born in $\rightarrow$ citizen of). 
The EOP contains a wide range of knowledge resources, 
including lexical and syntactic resources. 
Part of them are mannually grabbed from dictionaries, while others are automatically learned. 
The EOP platform includes three different approaches to RTE: 
i) an EDA based on transformations between $T$ and $H$; 
ii) an EDA based on edit distance algorithms; and 
iii) a classification based EDA using features extracted from $T$ and $H$.

Transformation-based EDA applies a sequence of transformations on $T$ with the goal of making it identical to $H$. 
Consider the following 
example where the text is ”The boy was located by the police” and the hypothesis is ''The child was found by the police``. 
Two transformations: $ boy\rightarrow child$ and $located \rightarrow found$ do the job.

Edit distance EDA involves using algorithms casting textual entailment as the problem of mapping the whole content of $T$ into the content of $H$.
Mappings are performed as sequences of editing operations (i.e., insertion, deletion and substitution) on text portions needed to transform $T$ into $H$, where each edit operation has an associated cost. 
The underlying intuition is that the probability of an entailment relation
between $T$ and $H$ is related to the distance between them.

Classification based EDA uses a maximum entropy classifier to combine the outcomes of several scoring functions and to learn a classification model for recognizing entailment. 
The scoring functions extract a number of features at various linguistic levels (bag-of-words, syntactic 
dependencies, semantic dependencies, named entities)~\cite{magnini2014excitement} 

MaxEntClassificationEDA is an Entailment Decision Algorithm (EDA) based on a prototype system called Textual Inference Engine (TIE).
Results for the three EDAs included in the EOP platform are reported in Table~\ref{tab:eop}. 
Each line represents an EDA, the language and the dataset on which the EDA was evaluated.

\begin{table}
\begin{center}
\begin{footnotesize}
\caption{Result for the entailment decision algorithm.}
\label{tab:eop}
\begin{tabular}{|p{5cm}|l|l|}
\hline
{\it EDA} & {\it Accuracy}  \\
\hline
Transformation-based English RTE-3 & 67.13"\%" \\ \hline 
Transformation-based English RTE-6 & 49.55"\%" \\ \hline 
Edit-Distance English RTE-3 & 64.38"\%" \\ \hline 
Edit-Distance German RTE-3 & 59.88"\%" \\ \hline 
Edit-Distance Italian RTE-3 & 63.50"\%" \\ \hline 
Classification-based English RTE-3 & 65.25"\%" \\ \hline 
Classification-based German RTE-3 & 63.75"\%"   \\ \hline 
Median of RTE-3 (English) submissions & 61.75"\%" \\ \hline 
Median of RTE-6 (English) submissions & 33.72"\%" \\ \hline 
\end{tabular}
\end{footnotesize}
\end{center}
\end{table}

\subsection{Natural language processing with GATE.}
For natural language procesing we use GATE (General Architecture For Text Engineering).
In GATE the logic is arranged in modules that are called pipelines. 
GATE  contains an information extraction pipeline called ANNIE composed of several components: Tokenizer, Gazetteer List,Sentence Splitter, POS Tagger, Semantic Tagger that annotates entities such as Person, Organization, Location, and an Orthographic  Co-reference that adds identity relations between the entities annotated by the Semantic Tagger. 
The tokeniser splits the text into very simple tokens such as numbers, punctuation and words of diﬀerent types.
The role of the gazetteer is to identify entity names in the text based on lists. 

\section{Running scenario}
\label{sec:scenario}

Consider the essay in example~\ref{ex:scenario}.

\begin{example}[Sample essay in the chemical domain]

$ $

\begin{footnotesize}
{\it ``All the matter in the universe is composed of the atoms of more than 100 different chemical elements, which are found both in pure form 
and combined in chemical compounds.First,a sample of any given pure element is composed only of the atoms characteristic of that element,
and the atoms of each element are unique. For example, the atoms that constitute carbon are different from those that make up iron, which
are in turn different from those of gold. Every element is designated by a unique symbol consisting of one or more letters arising from either
the current element name or its original Latin name. For example, the symbols for carbon ,hydrogen, and oxygen are simply C, H, and O, 
respectively. The symbol for iron is Fe, from its original Latin name ferrum. On the other hand, the chemical compound is any substance
composed of identical molecules consisting of atoms of two or more chemical elements.
The fundamental principle of the science of chemistry is that the atoms of different elements can combine with one another to form chemical 
compounds.Methane, for example, which is formed from the elements carbon and hydrogen in the ratio four hydrogen atoms for each carbon atom, 
is known to contain distinct CH4 molecules. The formula of a compound—such as CH4—indicates the types of atoms present, with subscripts
representing the relative numbers of atoms. 
Water, which is a chemical compound of hydrogen and oxygen in the ratio two hydrogen atoms for every oxygen atom, contains H2O molecules.
Sodium chloride is a chemical compound formed from sodium (Na) and chlorine (Cl) in a 1:1 ratio. Although the formula for sodium chloride is 
NaCl, the compound does not contain actual NaCl molecules. Rather, it contains equal numbers of sodium ions with a charge of positive one (Na+) 
and chloride ions with a charge of negative one (Cl−).The substances mentioned above exemplify the two basic types of chemical compounds: 
molecular (covalent) and ionic. Sodium chloride, on the other hand, contains ions so we can say that it is an ionic compound.''}
\end{footnotesize}
\label{ex:scenario}
\end{example}


\begin{figure}
\begin{footnotesize}
AcylBromide~\ensuremath{\equiv}~AcylHalide~\ensuremath{\sqcap}~\ensuremath{\exists}hasPart.AcylBromideGroup\\
AcylChloride~\ensuremath{\equiv}~AcylHalide~\ensuremath{\sqcap}~\ensuremath{\exists}hasPart.AcylChlorideGroup\\
AcylCompound~\ensuremath{\equiv}~OrganicCompound~\ensuremath{\sqcap}~\ensuremath{\exists}hasPart.AcylGroup\\
AcylFluoride~\ensuremath{\equiv}~AcylHalide~\ensuremath{\sqcap}~\ensuremath{\exists}hasPart.AcylFluorideGroup\\
AcylHalide~\ensuremath{\equiv}~OrganicCompound~\ensuremath{\sqcap}~\ensuremath{\exists}hasPart.AcylHalideGroup\\
AcylIodide~\ensuremath{\equiv}~AcylHalide~\ensuremath{\sqcap}~\ensuremath{\exists}hasPart.AcylIodideGroup\\
Alcohol~\ensuremath{\equiv}~OrganicCompound~\ensuremath{\sqcap}~\ensuremath{\exists}hasPart.HydroxylGroup\\
Aldehyde~\ensuremath{\equiv}~OrganicCompound~\ensuremath{\sqcap}~\ensuremath{\exists}hasPart.AldehydeGroup\\
Amide~\ensuremath{\equiv}~OrganicCompound~\ensuremath{\sqcap}~\ensuremath{\exists}hasPart.AmideGroup\\
Amine~\ensuremath{\equiv}~OrganicCompound~\ensuremath{\sqcap}~\ensuremath{\exists}hasPart.AmineGroup\\
Atom~\ensuremath{\sqsubseteq}~\ensuremath{\lnot}OrganicCompound\\
CarbonylCompound~\ensuremath{\equiv}~OrganicCompound~\ensuremath{\sqcap}~\ensuremath{\exists}hasPart.CarbonylGroup\\
CarboxylicAcid~\ensuremath{\equiv}~OrganicCompound~\ensuremath{\sqcap}~\ensuremath{\exists}hasPart.CarboxylicAcidGroup\\
Ester~\ensuremath{\equiv}~OrganicCompound~\ensuremath{\sqcap}~\ensuremath{\exists}hasPart.EsterGroup\\
Ether~\ensuremath{\equiv}~OrganicCompound~\ensuremath{\sqcap}~\ensuremath{\exists}hasPart.EtherGroup\\
HalogenCompound~\ensuremath{\equiv}~OrganicCompound~\ensuremath{\sqcap}~\ensuremath{\exists}hasPart.HalogenAtom\\
Hydrocarbon~\ensuremath{\equiv}~OrganicCompound~\ensuremath{\sqcap}~\ensuremath{\exists}hasPart.CarbonAtom~\ensuremath{\sqcap}~\ensuremath{\exists}~hasPart~HydrogenAtom~\ensuremath{\sqcap}~\ensuremath{\forall}~hasPart~(CarbonAtom~\ensuremath{\sqcup}~HydrogenAtom)\\
Imine~\ensuremath{\equiv}~OrganicCompound~\ensuremath{\sqcap}~\ensuremath{\exists}hasPart.ImineGroup\\
Ketone~\ensuremath{\equiv}~OrganicCompound~\ensuremath{\sqcap}~\ensuremath{\exists}hasPart.KetoneGroup\\
OrganicCompound~\ensuremath{\equiv}~Compound~\ensuremath{\sqcap}~\ensuremath{\exists}hasPart.CarbonGroup\\
OrganicCompound~\ensuremath{\sqsubseteq}~\ensuremath{\lnot}Atom\\
OrganicSulfurCompound~\ensuremath{\equiv}~OrganicCompound~\ensuremath{\sqcap}~\ensuremath{\exists}hasPart.OrganicSulfurG\\
PrimaryAmine~\ensuremath{\equiv}~OrganicCompound~\ensuremath{\sqcap}~\ensuremath{\exists}hasPart.PrimaryAmineGroup\\
SecondaryAmine~\ensuremath{\equiv}~OrganicCompound~\ensuremath{\sqcap}~\ensuremath{\exists}hasPart.SecondaryAmineGroup\\
TertiaryAmine~\ensuremath{\equiv}~OrganicCompound~\ensuremath{\sqcap}~\ensuremath{\exists}hasPart.TertiaryAmineGroup\\
\end{footnotesize}
\caption{Part of the chemical ontology.}
\label{fig:onto}
\end{figure}


\begin{table}
\begin{footnotesize}
\caption{Sample of textual entailment results.}
\label{tab:results}
\begin{tabular}{|p{5.5cm}|l|l|}
\hline
{\it Hypothesis} & {\it Confidence}  \\
\hline
Every AcylBromide is an AcylHalide that hasPart an AcylBromideGroup & 0.9999997639035405\\ \hline 
Every Alcohol is an OrganicCompound that hasPart a HydroxylGroup & 0.9999999948280452\\ \hline 
Every Aldehyde is an OrganicCompound that hasPart an AldehydeGroup & 0.9999997639035405\\ \hline 
Every Amide is an OrganicCompound that hasPart an AmideGroup & 0.9999997639035405\\ \hline 
Every OrganicSulfurCompound is an OrganicCompound that hasPart an OrganicSulfurGroup & 0.9999997639035405\\ \hline 
No Atom is an OrganicCompound & 0.9999991153658677\\ \hline 
\end{tabular}
\end{footnotesize}
\end{table}

We run a use case scenario based on the essay in example~\ref{ex:scenario}.
Firstly, we load the domain knowledge, that is the ontology in chemical domain shown in Fig.~\ref{fig:onto}.
Secondly, we generate the list of hypothesis based on the ontology, like in Fig.\ref{fig:verbalizer}.
Thirdly, we select a specific number of hypothesis, which will be used by the component of textual entailment.
Finnaly, we load the essay, and we run the component for textual entailment.  
The system runs each hypothesis on the essay and returns a confidence value in [0..10] for each hypothesis.
Performing these steps in the essay in example~\ref{ex:scenario} on six hypothesis, we obtain the results in Table~\ref{tab:results}.
These confidece values are used to compute the grade.


Consider a set of 10 essay to assess in the chemical domain. 
Assume that the domain ontology have been already generated by TextToOnto or available from various ontology repositories.
The professor has the following four tasks: 
\begin{enumerate}
 \item Load the domain ontology (i.e. organic\_compound.owl);
 \item Generate all the hypothesis in natural language from that ontology (based on OWLNatural);  
\item  Select the hypothesis against which the essay should be verified; 
\item Load the essay to be check if entails the selected hypothesis (based on textual entailment).
\end{enumerate}

\begin{table}
\begin{footnotesize}
\caption{Grading 10 essays with 10 hypothesis ($H_{10}$) and 20 hypothesis ($H_{10}$).}
\begin{tabular}{|l|l|l|l|l|}\hline
{\it Essay} & $Grade(H_{10})$ & $Time (H_{10})$ & $Grade(H_{20})$ & $Time (H_{20})$ \\ \hline
1 & 8.2 & 2.5 & 8.5 & 4  \\ \hline
2 & 8.5& 1.9 &9 & 3.8 \\ \hline
3 & 8& 2.75& 8.5& 4.5 \\ \hline
4 & 7& 2.8& 8& 4.5\\ \hline
5 & 7.5& 2.25& 8& 3.8\\ \hline
6 & 7.3& 2& 7.8& 3.75\\ \hline
7 & 7.5& 2.5& 8& 4\\ \hline
8 & 8& 2.9&8.3 &3.5 \\ \hline
9 & 8& 2.3& 8& 3.75\\ \hline
10 & 7.5& 2.5& 7.8&4 \\ \hline
\end{tabular}
\end{footnotesize}
\label{tab:eseuri}
\end{table} 


Table~\ref{tab:eseuri} shows the assessment when the user selects different number of hypothesis.  
The execution time varies between 1.5 and 3 minutes for 10 hypothesis, and between 3.5 and 4.5 minutes in case of 20 hypothesis. 
Table~\ref{tab:comp} shows the results obtained after the comparison with the similar system PaperRater, and manual evaluation. 

\begin{table}
\begin{center}
\begin{footnotesize}
\caption{Comparison with similar systems and human evaluation.}
\label{tab:comp}
\begin{tabular}{|l|l|l|l|l|}\hline
{\it Essay ID} & Our system & PaperRater & Manual evaluation  \\ \hline
1 & 8.5 & 8.5 & 9 \\ \hline
2 & 9 & 8.8 & 9.5 \\ \hline
3 & 8 & 8.1& 8.5 \\ \hline
4 & 8.3 & 9 & 9.5 \\ \hline
5 & 8 & 8.9 & 9 \\ \hline
6 & 8.5 & 8.3 & 8 \\ \hline
7 & 8 & 8.1 & 7.5 \\ \hline
8 & 8.5 & 8.2 & 8 \\ \hline
9 & 8 & 8.3 & 9 \\ \hline
10 & 7.5 & 7.8 & 7  \\ \hline
\end{tabular}
\end{footnotesize}
\end{center}
\end{table} 

\section{Discussion and related work}
\label{sec:related}


The current methods of essay scoring can be categorized into two classes: holistic scoring and rubric-based scoring.
In holistic scoring, the essay is assessed and a single score selected from a predefined score range is assigned as an overall score~\cite{Naplan4}. 
In the analytical or rubric-based scoring method, essays are assessed on the basis of a certain set of well-defined features~\cite{Naplan4}. 
Each feature has a scale associated with it and the final score awarded to the essay is the sum of scores of all the essay rubrics/features.



The National Assessment Program Literacy and Numeracy (NAPLAN)~\cite{fazal2013innovative} rubric of persuasive essay grading lists a set of criteria for marking persuasive writing.
The Spelling Mark algorithm is developed to formalise the NAPLAN rubric for marking spelling based on common heuristics and rules of the English language. 
The first step is to obtain the total number of words in the essay and the number of spelling errors in the essay. 
Then, each word is categorized based on the difficulty level into one of four classes: simple, common, difficult or challenging, while the number of correct and incorrect words in each category is counted. 
The final step in the algorithm is to assign the spelling mark according to the set of rules~\cite{fazal2013innovative}.

PaperRater (https://www.paperrater.com/) is an automated proofreading system that combines NLP,
machine learning, information retrieval and data mining to help students write better. 
PaperRater is also used by schools and universities in over 46 countries to check for plagiarism. 
The system has a core NLP engine using statistical and rules
to extract language features from essays and translate that into statistical models. 
The three major features are: spell checker, grammar checker, and plagiarism checker. 
The tool also has a vocabulary builder tool designed to help students learn proper usage of more sophisticated words.

A complementary line of research is given by argumentative writing support systems~\cite{letia2012arguing}. 
Assisting students in essays with structured and sound arguments is an important educational goal~\cite{scheuer2010computer}. 
Hence, various argumentative support systems has been applied in collaborative educational environments~\cite{belgiorno2008face,pinkwart2012educational}.
Persuasive essays are extensively studied in the context of automated essay grading. 
Since argument structure is crucial for evaluating essay quality,~\cite{stab2014identifying} identifies the argumentative discourse structure by means of discourse marking.
The goal is to model argument components as well as argumentative relations that constitute the argumentative discourse  structure in persuasive essays. 
The annotation scheme includes three argument components (major claim, claim and premise) and two argumentative relations (support and attack)~\cite{stab2014identifying}.
The legal educational system LARGO~\cite{pinkwart2008re} uses an ontology containing the concepts ``test'', ``hypothetical'', and ``fact situation'' and roles such as ``distinction'' and ``modification'', while NLP based queries are used to interrogate biomedical data in~\cite{anca2015}. 
The system Convince Me~\cite{schank1995improved} employs more scientific-focused 
primitives such as “hypothesis” and “data” elements with “explain” and “contradict” links. 
Other systems such as Rationale~\cite{van2007rationale} provide more expansive primitive sets that allow users to construct arguments in different domains.
Our work fits in this context by using natural language processing for argument mining. 
The automatic grading mechanism could benefit from a system able to identify arguments in a student essay.


\section{Conclusion}
\label{sec:con}

We developed a NLP tool for automatically essay grading in different domains. 
We enacted textual entailment to compare the text written by a student with the requirments of a human evaluator. 
These requirments are generated in natural language from a given domain ontology. 
The main benefit is that the user can select different ontologies for processing text in various domains. 
However, the confidence in the assessment depends on the precision of the textual entailment method that relies on domain datasets for training. 

In line with~\cite{manap2012comparison}, we currently aim to assess the confidence in the system by performing more comparisons with humans that evaluate essays. 

\section*{Acknowledgments}
We thank the reviewers for valuable comments. This research was supported by the Technical University of Cluj-
Napoca, Romania, through the internal research project GREEN-VANETS.

\bibliographystyle{IEEEtran}
\bibliography{cop} 

\end{document}